\documentclass[sigconf]{acmart}
\usepackage{multirow}
\pdfoutput=1 
\usepackage{colortbl}
\usepackage{bbding}
\AtBeginDocument{%
  \providecommand\BibTeX{{%
    \normalfont B\kern-0.5em{\scshape i\kern-0.25em b}\kern-0.8em\TeX}}}

\setcopyright{acmlicensed}
\copyrightyear{2018}
\acmYear{2018}
\acmDOI{XXXXXXX.XXXXXXX}

\acmConference[MM'24]{Make sure to enter the correct
  conference title from your rights confirmation email}{October 28 - November 1,
  2024}{Melbourne, Australia.}
\acmISBN{978-1-4503-XXXX-X/18/06}

\begin{document}

\title{LIR: A Lightweight Baseline for Image Restoration}


\author{Dongqi Fan}
\affiliation{%
  \institution{University of Electronic Science and Technology of China}
  % \streetaddress{1 Th{\o}rv{\"a}ld Circle}
  \city{Chengdu}
  \country{China}}
\email{dongqifan@std.uestc.edu.cn}

\author{Ting Yue}
\affiliation{%
  \institution{University of Electronic Science and Technology of China}
  % \streetaddress{1 Th{\o}rv{\"a}ld Circle}
  \city{Chengdu}
  \country{China}}
\email{tingyue@std.uestc.edu.cn}

\author{Xin Zhao}
\affiliation{%
  \institution{University of Electronic Science and Technology of China}
  % \streetaddress{1 Th{\o}rv{\"a}ld Circle}
  \city{Chengdu}
  \country{China}}
\email{xinzhao@std.uestc.edu.cn}

\author{Renjing Xu}
\affiliation{%
  \institution{The Hong Kong University of Science and Technology (Guangzhou)}
  % \streetaddress{1 Th{\o}rv{\"a}ld Circle}
  \city{Guangzhou}
  \country{China}}
\email{renjingxu@hkust-gz.edu.cn}

\author{Liang Chang}
\affiliation{%
  \institution{University of Electronic Science and Technology of China}
  % \streetaddress{1 Th{\o}rv{\"a}ld Circle}
  \city{Chengdu}
  \country{China}}
\email{liangchang@uestc.edu.cn}
\renewcommand{\shortauthors}{author name and author name, et al.}

\begin{abstract}
Recently, there have been significant advancements in Image Restoration based on CNN and transformer. However, the inherent characteristics of the Image Restoration task are often overlooked in many works. They, instead, tend to focus on the basic block design and stack numerous such blocks to the model, leading to parameters redundant and computations unnecessary. Thus, the efficiency of the image restoration is hindered. In this paper, we propose a Lightweight Baseline network for Image Restoration called LIR to efficiently restore the image and remove degradations. First of all, through an ingenious structural design, LIR removes the degradations existing in the local and global residual connections that are ignored by modern networks. Then, a Lightweight Adaptive Attention (LAA) Block is introduced which is mainly composed of proposed Adaptive Filters and Attention Blocks. The proposed Adaptive Filter is used to adaptively extract high-frequency information and enhance object contours in various IR tasks, and Attention Block involves a novel Patch Attention module to approximate the self-attention part of the transformer. On the deraining task, our LIR achieves the state-of-the-art Structure Similarity Index Measure (SSIM) and comparable performance to state-of-the-art models on Peak Signal-to-Noise Ratio (PSNR). For denoising, dehazing, and deblurring tasks, LIR also achieves a comparable performance to state-of-the-art models with a parameter size of about 30\%. In addition, it is worth noting that our LIR produces better visual results that are more in line with the human aesthetic. The code is available at: \href{https://github.com/Dongqi-Fan/LIR}{https://github.com/Dongqi-Fan/LIR}
\end{abstract}



% \keywords{Image Restoration; \and Patch Attention; \and Adaptive Filter; \and Residual Connection; \and Lightweight}



\maketitle

\begin{figure*}[htbp]
\includegraphics[width=\linewidth]{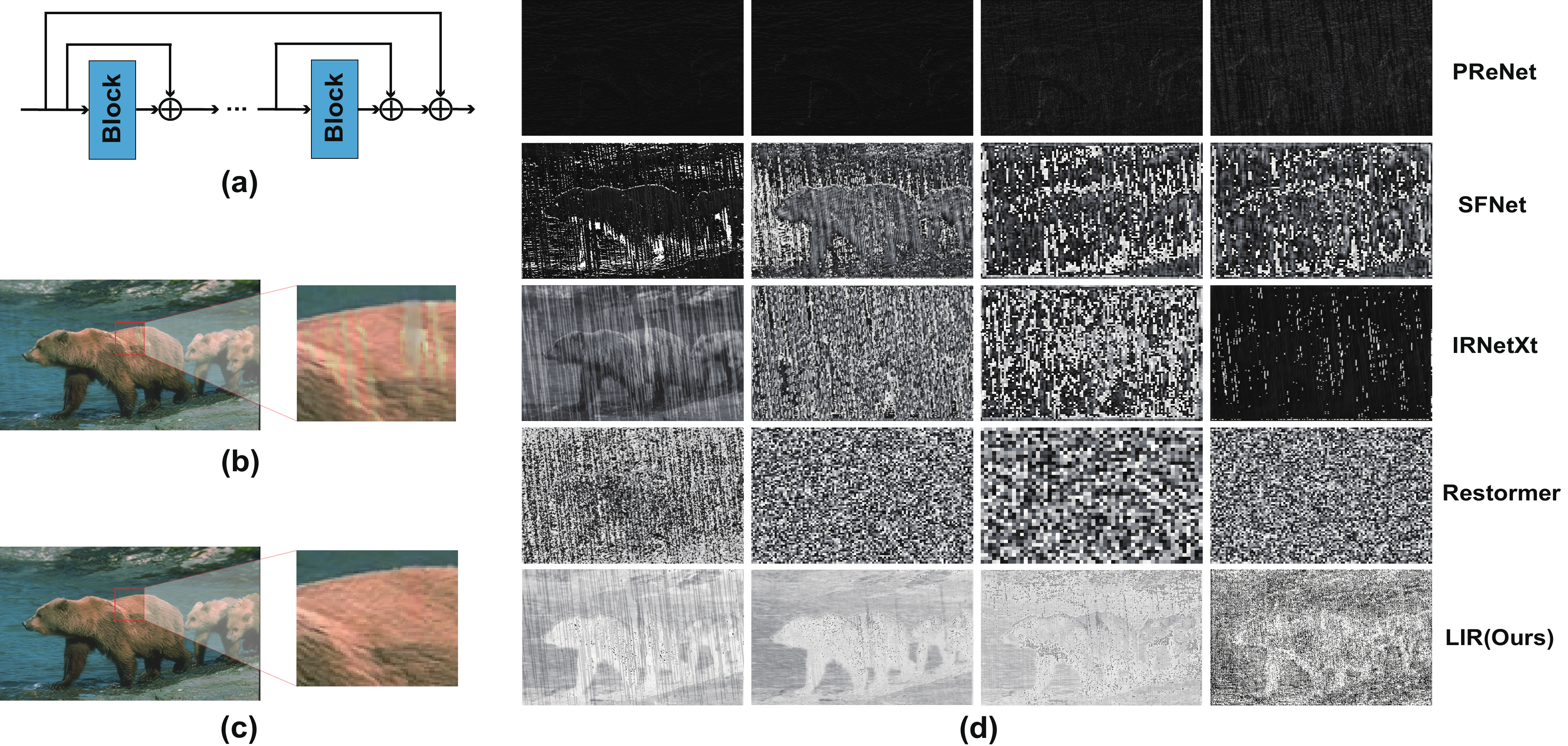}
\caption{Visualization experiment on the deraining task. (a) The paradigm of the most modern networks that incorporate both local and global residual connections. (b) The restored image output by the state-of-the-art model, Restormer \cite{34zamir2022restormer}. It still exhibits visible traces of rain. (c) The restored image output by our LIR. It is cleaner and solves the problem of residual connections carrying degradation. (d) The visualization of intermediate feature maps of PReNet \cite{55ren2019progressive}, SFNet \cite{51cui2022selective}, IRNeXt \cite{cui2023irnext}, Restormer, and LIR.}
\label{fig:fig1}
\end{figure*}

\section{Introduction}
\label{sec:Introduction}
Image Restoration (IR) is a fundamental computer vision task aiming to reconstruct a degraded image into a clean one, including deblurring, deraining, etc. In recent years, we have witnessed the great process of deep learning methods 
\cite{29liang2021swinir,30zamir2021multi,31tu2022maxim,32luo2020latticenet,33poirier2023robust,12wang2022uformer,18zhang2023kbnet} in IR. However, many existing works tend to focus on designing the basic block and inefficiently stacking these blocks in a model, without considering the specific characteristics of the IR task. This approach often leads to parameter redundancy and computations unnecessary. Instead, our goal is an efficient and lightweight network design for various IR tasks.

\begin{figure*}[tb]
\centering
\includegraphics[width=\linewidth]{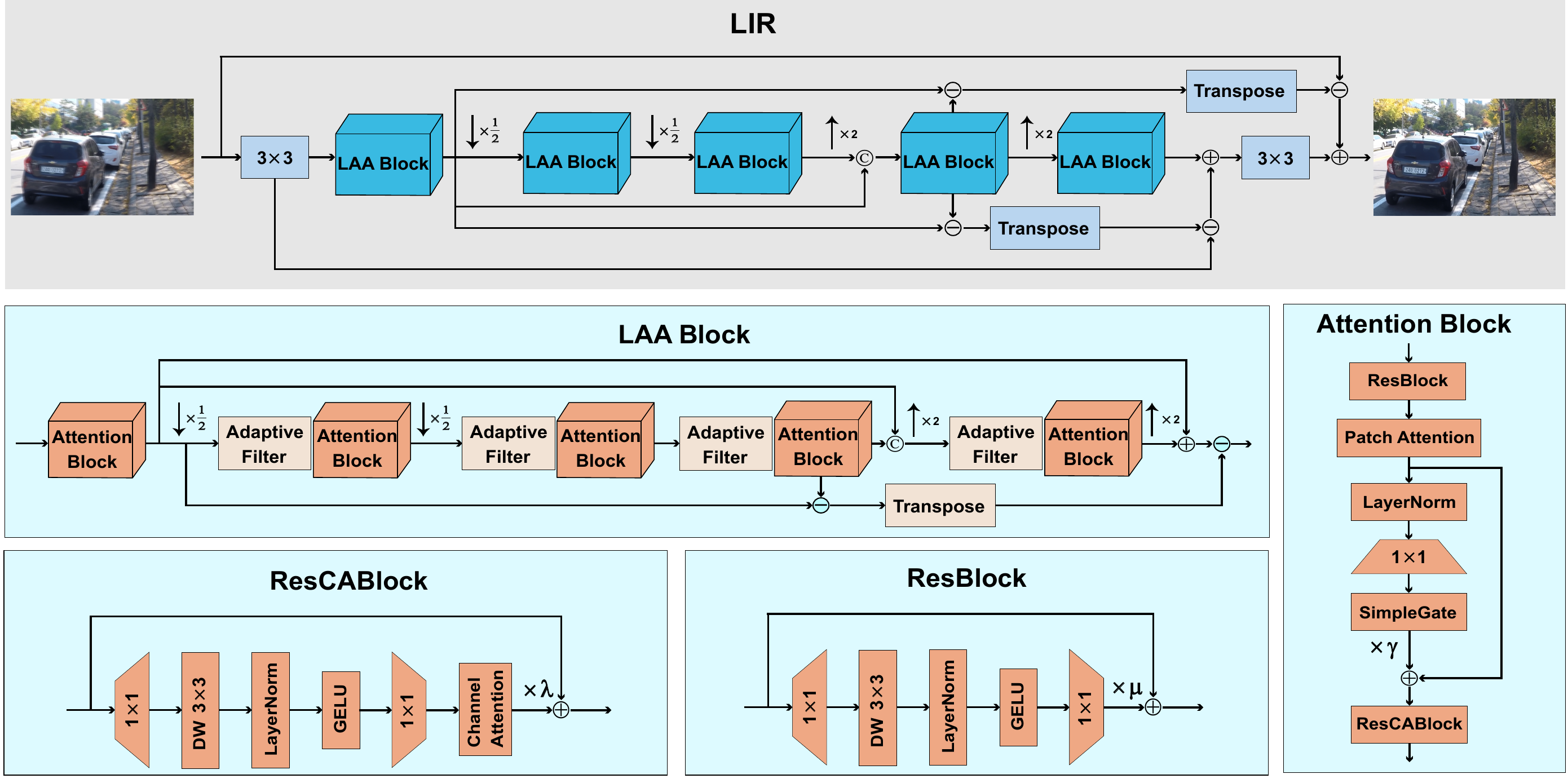}
\caption{The detailed structure of the proposed Lightweight Image Restoration (LIR) network, which is mainly stacked by Lightweight Adaptive Attention (LAA) Blocks. Thanks to the Adaptive Filter, Patch Attention, ResCABlock, and the ResBlock, the LAA Block is endowed with the ability to extract a variety of information efficiently.}     
\label{fig:fig2}
\end{figure*}

To address the issue of vanishing gradient and stabilize training, as shown in Figure~\ref{fig:fig1} (a), most networks usually incorporate local and global residual connections. However, these connections may propagate the degradation of the input throughout the entire network and introduce it into the reconstructed clean output. To validate our idea, we conduct visualization experiments on deraining task using Restormer \cite{34zamir2022restormer}, SFNet \cite{51cui2022selective}, IRNeXt \cite{cui2023irnext}, and PReNet \cite{55ren2019progressive}. We select intermediate feature maps near the input and output of the networks for visualization because the semantics of the middle layer are too abstract. Results are shown in Figure~\ref{fig:fig1} (d). Notably, it is evident that each feature map of these networks and the final output of the state-of-the-art model (Figure~\ref{fig:fig1} (b)), Restormer, all contain severe degradation (rain). To overcome this issue, we use an ingenious structural design, using transpose branches in the LIR and LAA (Figure~\ref{fig:fig2}), that effectively eliminates the degradation in the local and global residual connections. The intermediate feature maps of our LIR (Figure~\ref{fig:fig1} (d)) show better visual performance compared with the rest of the networks, and the restored image output by our LIR (Figure~\ref{fig:fig1} (c)), which is more cleaner than the Restormer (Figure~\ref{fig:fig1} (b)). More visual comparison of our LIR and these networks see Supplementary Materials.

Moreover, we observe another issue in the visualization experiment. As shown Figure~\ref{fig:fig1} (d)), the intermediate feature maps of the PReNet, SFNet, IRNetXt, and Resormer not only exhibit blurred contours but also contain a considerable amount of noise. The more important is that the high-frequency information has been lost. These observations indicate that although many state-of-the-art networks have achieved satisfactory results on the metric (PSNR), the restoration process of the image is highly inefficient and missing to address the core issues of image restoration, thus resulting in poor visual quality. Based on this insight, we propose the Adaptive Filter (Figure~\ref{fig:fig3}) (a component of the LAA) as a solution to adaptively extract high-frequency information, enhance object contours, and remove degradation in various IR tasks.

Transformer-based models have demonstrated impressive performance in various tasks, both at high-level \cite{35dosovitskiy2020image,36dong2022cswin,37liu2021swin,38touvron2021training} and low-level \cite{28chen2022activating,5chen2021pre,12wang2022uformer,13tsai2022stripformer,14li2023efficient}. However, there exist different opinions. \cite{1dong2021attention} argues that attention is not all you need, emphasizing the importance of skip connections and Multi-layer Perceptrons (MLPs). In addition, \cite{4liu2021pay,6tolstikhin2021mlp,7touvron2022resmlp} share a similar viewpoint that highlights the potential of MLPs beyond transformers. On the other hand, \cite{8trockman2022patches} suggests that CNN can achieve comparable results to transformers by leveraging the attention mechanism and patch embedding strategy. Thus, excessive computations can be avoided. Furthermore, \cite{2guo2023visual,9liu2022convnet,3ding2022scaling,10zhou2022efficient} take advantage of the attention mechanism in the CNN along with large convolution kernels to capture global information, instead of the self-attention among the transformer. Satisfactory results are achieved and excessive computations are avoided. Inspired by these perspectives, through leveraging the patch embedding strategy and MLP, we propose the Patch Attention module (Figure~\ref{fig:fig4}) (an component of the Attention Block) that aims to capture global information of the features and keeps parameters and computations friendly.

In this paper, we propose a Lightweight Image Restoration network called LIR (Figure~\ref{fig:fig2}) to efficiently remove degradation and adapt to different IR tasks. The key component of LIR is the novel Lightweight Adaptive Attention Block (LAA) consisting of several proposed Adaptive Filters and Attention Blocks. Where the Attention Block is formed by proposed Patch Attention module (Figure~\ref{fig:fig4}), ResCABlock, and ResBlock (details can be found in Figure~\ref{fig:fig2}). Thanks to the Adaptive Filter and Attention Block, the LAA Block is capable of adaptively sharpening contours, removing degradation, and capturing global information in various Image Restoration scenes in a computation-friendly manner. Furthermore, we address the issue of degradation existing in the local and global residual connections by subtracting the output of the later component from the output of the first component in the LIR and LAA Block. This subtraction allows us to obtain the degradation, which is then expanded through the Transpose operation. After that, we subtract this expanded degradation from the residual connection to obtain a clean residual connection. Experiments demonstrate that our LIR achieves the state-of-the-art SSIM and comparable PSNR to state-of-the-art models on the deraining task, and comparable performance to state-of-the-art models with a parameter size of about 30\% on the denoising, dehazing, and deblurring tasks.

Our contributions can be summarized as follows:

\begin{itemize}
\item Through an ingenious structural design, the problem that local and global residual connections bring degradation is addressed, which is overlooked by modern networks.

\item Given the problem of the loss of high-frequency information observed in the visualization experiments, we propose the Adaptive Filter (Figure~\ref{fig:fig3}) as a solution to adaptively extract high-frequency information and enhance object contours.

\item To give LIR the ability to extract global information, we propose the Patch Attention module (Figure~\ref{fig:fig4}), which is more computation-friendly than the self-attention of the transformer.

\item Based on Adaptive Filter, Patch Attention, ResCABlock, and ResBlock, we designed a lightweight and powerful block: LAA. It can make the LIR achieve comparable performance to state-of-the-art networks and better visual effects with smaller parameters and computations.

\end{itemize}

\section{Related Work}
\label{sec:Related_Work}

\subsection{Image Restoration}
\label{Image_Restoration}
In recent years, IR based on deep learning methods \cite{40du2020learning,19zhang2023ingredient,16xie2021finding,41li2020all,42zhang2022practical,43huang2023contrastive} have shown promising results. The transformer architectures achieve adaptive spatial aggregation at the cost of heavy computation. To address this challenge, a Kernel Basis Attention (KBA) module has been proposed by KBNet \cite{18zhang2023kbnet} in the CNN. It adaptively aggregates spatial neighborhood information in a kernel-wise manner, thereby reducing computation. ADMS \cite{17park2023all} proposes a Filter Attribution method based on Integral Gradient (FAIG) \cite{16xie2021finding} to identify the contributions of the filters to remove specific degradations. By masking filters with minimal impact on degradation removal, the network's efficiency in removing degradations is improved. IDR \cite{19zhang2023ingredient} proposes a learnable Principal Component Analysis (PCA) and views various IR tasks as multi-task learning methods to obtain network priors and remove degradations. MPRNet \cite{30zamir2021multi} introduces the Multi-Stage Progressive Restoration method, which is based on the U-Net architecture and follows a multi-input multi-output style. In addition, the contrastive learning paradigm is utilized by AirNet \cite{54li2022all} to train the network to deal with different unknown degradations. Through the prompting of the contrastive learning paradigm, the latent features in the network gradually away from the degradation space and close to the clean space. The above works either involves a large number of parameters or a complex training process. Instead, our emphasis is on lightweight and efficient design while ensuring good performance.

\subsection{Vision Transformer for Image Restoration}
\label{Vision_Transformer}
IPT \cite{5chen2021pre} is the first work that introduces ViT \cite{35dosovitskiy2020image} to low-level tasks to achieve image super-resolution, denoising, and deraining. The computational complexity of the self-attention part in ViT is $O(N^2D)$, where $N=H$$\times$$W$ represents the length of the patch sequence and $D$ represents the dimension of the sequence. Although vision transformers are capable of capturing long-range dependencies, the self-attention component imposes a significant computing burden. The computational complexity of the self-attention increases quadratically with the input size. To mitigate this issue, TransWeather \cite{11valanarasu2022transweather} introduces a reduction ratio factor R to control the complexity, resulting in a computational complexity of $O(N^2D/R)$. Uformer \cite{12wang2022uformer} and StripFormer \cite{13tsai2022stripformer}  proposes the window and stripe attention mechanisms respectively, which reduce the computational complexity to $O(NM^2D)$ and $O(D(H^2W+HW^2))$ by calculating dependencies in smaller windows and strips, where M represents the window size. To further reduce the computational complexity, \cite{15kong2023efficient} estimates the attention map that is output by the self-attention part using an element-wise product operation in the frequency domain, and GRL \cite{14li2023efficient} proposes an anchor stripe attention mechanism to futher narrow down the computational range of the slef-attention. The computational complexities are reduced to $O(NlogND)$ and $O(NMD)$ respectively, where M represents the size of feature map. However, the computation required for self-attention remains unaffordable. To address this issue, we propose a novel Patch Attention module to approximate the self-attention part by simplifying two matrix multiplications into one element-wise multiplication and removing the softmax operation for computational savings.

\section{Method}
\label{sec:method}

\begin{figure*}[tb]
\centering
\includegraphics[width=0.9\linewidth]{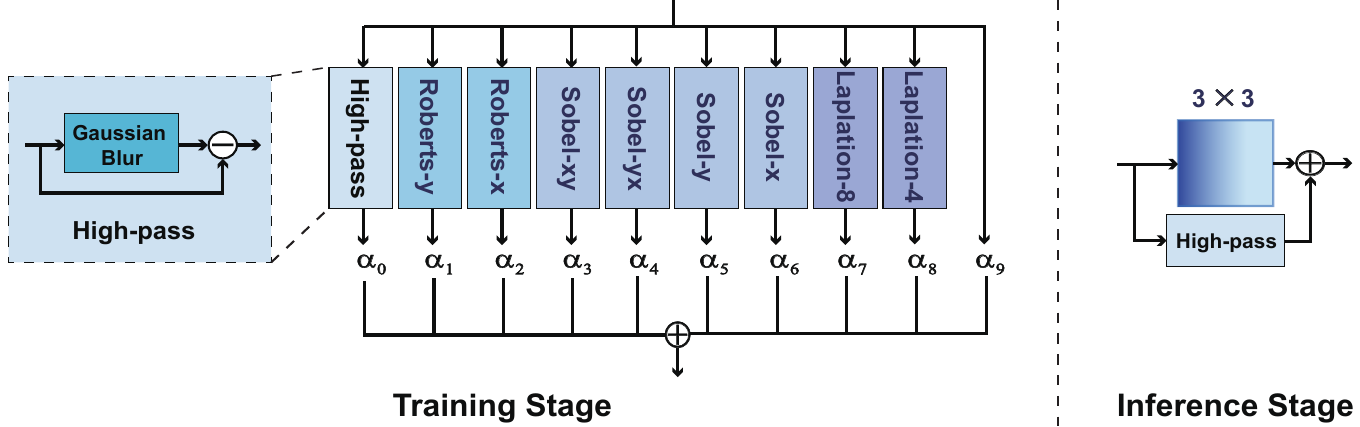}
\caption{The detailed architecture of the proposed Adaptive Filter. It is a reparameterization style that multiple convolutions are reparameterized into one during inference after the training.}  
\label{fig:fig3}
\end{figure*}

\begin{figure*}[tb]
\centering
\includegraphics[width=\linewidth]{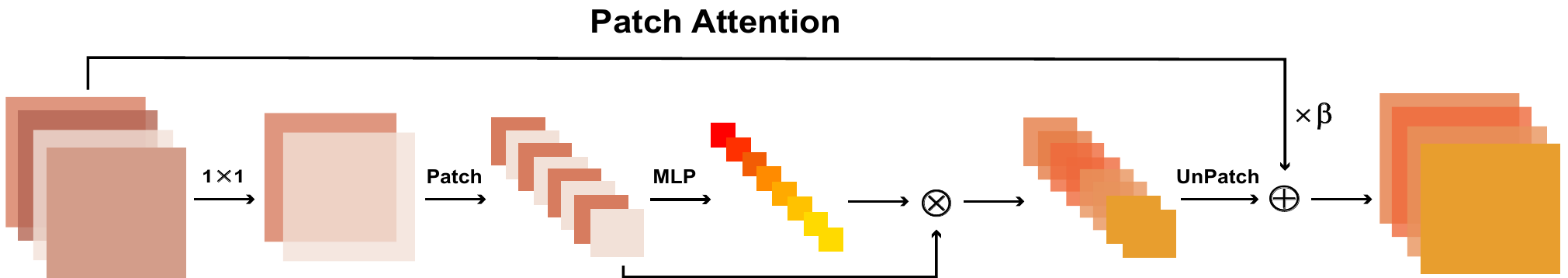}
\caption{The flow of the proposed Patch Attention.}     
\label{fig:fig4}
\end{figure*}

\subsection{Lightweight Image Restoration}
\label{LIR}
The overall structure of LIR is illustrated in Figure~\ref{fig:fig2}, which primarily consists of Lightweight Adaptive Attention (LAA) Blocks. Both LIR and LAA Block adopt a down-up style to reduce computation and save parameters, that is downsizing the feature map in the beginning and upsizing at the end. To preserve low-frequency information and ensure training is stable, we add two global residual connections and one local residual connection into LIR. For the global connections, one connects the input of the network to the output, while the other connects the head $3\times3$ Conv to the tail $3\times3$ Conv. For the local one, it connects the head and the tail.

As mentioned in Section~\ref{sec:Introduction}, these residual connections can bring degradation. To address this, we subtract the output of the fourth LAA Block from the output of the first LAA Block to obtain the corresponding degradation. We then amplify the degradation through the Transpose operation and subtract it from the residual connection to obtain a clean connection. The same approach is used to remove degradation in the local residual connection within the LAA Block, where we subtract the output of the fourth Attention Block from the output of the first Attention Block to obtain the corresponding degradation. The reason we adopt this method is that the feature map of the shallow layer contains more degradation compared to a deep layer. Thus, this design allows the network to dynamically perceive this difference between shallow and deep layers during the training process and inference stage. Now, the problem that local and global residual connections bring degradation is addressed. Next, we illustrate how we can achieve lightweight and efficiency through the Adaptive Filter and Attention Block in detail.

\subsection{Adaptive Filter}
\label{AF}
To achieve lightweight and efficiency, we first introduce the Adaptive Filter (Figure~\ref{fig:fig3}), inspired by Edge-oriented Convolution Block (ECB) \cite{48zhang2021edge}. The rich filters of different kinds allow the Adaptive Filter to enhance the object contour, remove degradation, and extract useful high-frequency information. Furthermore, the Adaptive Filter can be reparameterized as one $3\times3$ and one high-pass filter after training to achieve lightweight.

\begin{figure*}[t]
\centering
\includegraphics[width=\linewidth]{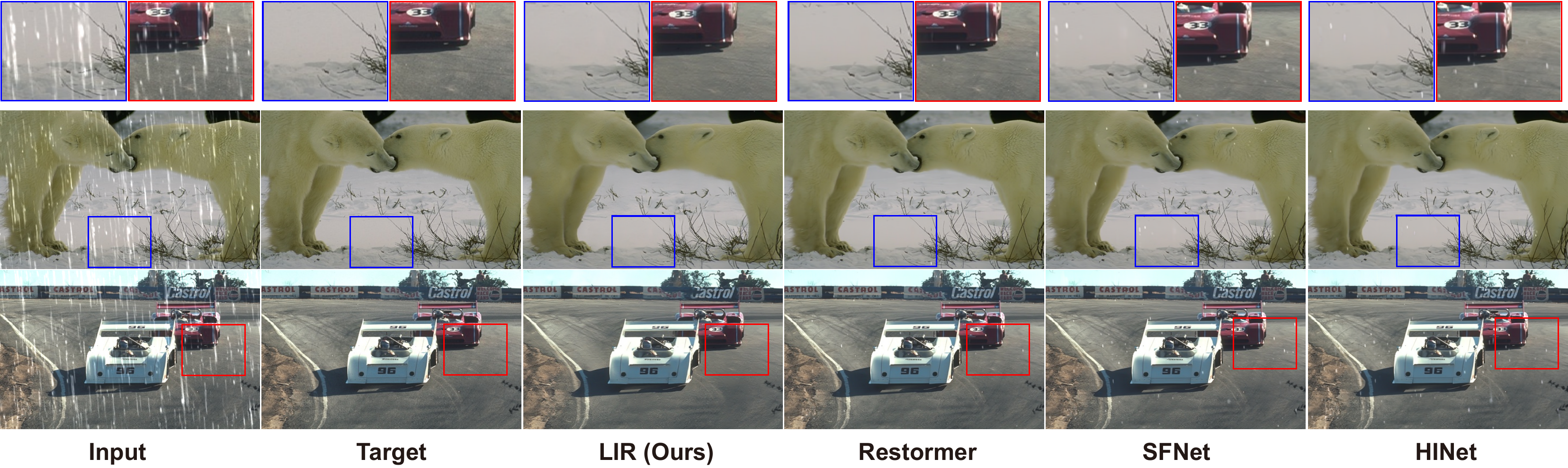}
\caption{The visual comparisons for LIR, Restormer \cite{34zamir2022restormer}, SFNet \cite{51cui2022selective}, and HINet \cite{50chen2021hinet}. LIR produces state-of-the-art visual results.}     
\label{fig:rain}
\end{figure*}

Our Adaptive Filter significantly outperforms ECB in three ways. Firstly, the ECB introduces more convolution operations during training leading to more training costs (i.e. parameters and computations). Instead, our Adaptive Filter only requires 10 training parameters resulting in smaller training costs. Secondly, the Adaptive Filter can be applied to various IR scenes and dynamically adjust the dependence on different filters by adjusting parameters $\alpha$. However, the ECB treats each branch equally and can not adapt to various IR scenes. Lastly, the Adaptive Filter offers a wider range of filters that can extract more types of useful information and is more robust to noisy scenes, while the ECB is brittle in rich degradation scenes.

Each filter in the Adaptive Filter has its own advantages and disadvantages. Thus, we use the learnable parameters $\alpha$ to adaptively maximize their advantages and minimize their disadvantages in different LAA blocks of the LIR. Specifically, in these filters, the \textbf{Roberts filter} is used to detect the pixel changes in the features by the differential method. It is effective in extracting prominent edges but fails to detect details. Where Roberts-x and Roberts-y represent the detection in the horizontal direction and vertical direction; The \textbf{Sobel filter} extracts edge information by approximating first-order differentials discretely. It has some resistance to noise and a good ability to detect details. However, the Sobel filter may cause edge thinning and false detection in scenes with a mass of noise; The \textbf{Laplacian filter} extracts edge information by approximating second-order differentials discretely. It can detect more detailed edges and textures than Sobel but is more sensitive to noise and may result in a bilateral effect on the edges. Where Laplacian-4 and Laplacian-8 represent 4 domains and 8 domains; The \textbf{high-pass filter} obtains high-frequency information by subtracting the features after Gaussian blur processing from the original features. This filter is more robust in noisy scenes compared to the above filters; Finally, we add a \textbf{residual connection} to the Adaptive Filter for stages where the features have scarce noise and a clear outline.

\subsection{Attention Block}
\label{AB}
While different types of filters in Adaptive Filter enable LAA blocks to extract rich information, Adaptive Filter alone is not enough, as we need to take full advantage of the rich information extracted by Adaptive Filter. Thus, a strong and efficient Attention Block is introduced, which consists of four components: ResBlock, Patch Attention (PA), SimpleGate, and ResCABlock. We first use the ResBlock to preprocess input features, followed by the extraction of global information using our PA module. The efficacy of the gate function has been demonstrated in prior research \cite{4liu2021pay,31tu2022maxim,46chen2022simple}. Here, we simply employ the SimpleGate module from \cite{46chen2022simple} to further extract dependencies between features. Finally, we leverage the ResCABlock to extract channel-wise information.

\begin{table}[htbp]
\caption{Image Deraining results evaluated by PSNR and SSIM on Rain100L dataset. }
\centering
\scalebox{0.85}{
\begin{tabular}{cccccc}
\toprule[1.5pt] 
\multirow{2}{*}{Method} & \multirow{2}{*}{Type} & \multicolumn{2}{c}{Rain100L \cite{rain100l}}               & \multirow{2}{*}{Params {[}M{]}} & \multirow{2}{*}{GFLOPs} \\ \cline{3-4}
                        &                       & PSNR                  & SSIM                 &                                 &                         \\ \midrule[0.9pt]
UMRL \cite{57yasarla2019uncertainty}                    & CNN                   & 29.18                & 0.923                & 0.98                            & 9.31                    \\
PReNet \cite{55ren2019progressive}                  & RNN                   & 32.44                & 0.950                & 0.17                            & 37.27                   \\
AirNet \cite{54li2022all}                  & CNN                   & 34.90                & 0.966                & 8.93                            & 169.47                  \\
MPRNet \cite{30zamir2021multi}                  & CNN                   & 36.40                & 0.965                & 20.13                           & 960.39                  \\
SPAIR \cite{53purohit2021spatially}                   & CNN                   & 36.93                & 0.969                & -                               & -                       \\
HINet \cite{50chen2021hinet}                  & CNN                   & 37.28                & 0.970                & 88.67                           & 95.92                   \\
IRNeXt \cite{cui2023irnext}                  & CNN                   & 38.14                & 0.972                & 5.46                           & 23.60                   \\
MAXIM-2S \cite{31tu2022maxim}               & MLP                   & 38.06                & 0.977                & 22.50                           & 80.13                   \\
SFNet \cite{51cui2022selective}                   & CNN                   & 38.21                & 0.974                & 13.27                           & 70.00                   \\
DRCNet \cite{49li2022drcnet}                  & CNN                   & 38.23                & 0.976                & 18.90                           & -                       \\
Restormer \cite{34zamir2022restormer}               & Transformer           & 38.99                & 0.978                & 26.13                           & 92.85                                     \\ \midrule[0.9pt]
LIR (Ours)              & CNN                   & 38.96                & 0.983                 & 7.31                            & 62.51                  \\ \bottomrule[1.5pt]
\multicolumn{1}{l}{}    & \multicolumn{1}{l}{}  & \multicolumn{1}{l}{} & \multicolumn{1}{l}{} & \multicolumn{1}{l}{}            & \multicolumn{1}{l}{}    \\
\multicolumn{1}{l}{}    & \multicolumn{1}{l}{}  & \multicolumn{1}{l}{} & \multicolumn{1}{l}{} & \multicolumn{1}{l}{}            & \multicolumn{1}{l}{}  
\end{tabular}}
\label{Rain100L}
\end{table}

\subsubsection{ResBlock and ResCABlock}
Both ResBlock and ResCABlock are up-down style residual blocks that include a learnable parameter in the trunk. The only difference between ResBlock and ResCABlock is that the latter has a channel attention module at the tail. The Channel Attention module \cite{44zhang2018image}, based on Squeeze-and-Excitation (SE) \cite{45hu2018squeeze}, has gained much popularity for its ability to extract channel-wise information. Many studies \cite{14li2023efficient,28chen2022activating,46chen2022simple,47zhao2020efficient} have directly utilized this module in their models. Following them, in our Attention Block, we also incorporate the Channel Attention module into the ResBlock to extract channel-wise information. In addition, the structure of ResBlock and ResCABlock are designed to be simple to meet lightweight and efficient requirements.

\begin{figure*}[htbp]
\centering
\includegraphics[width=\linewidth]{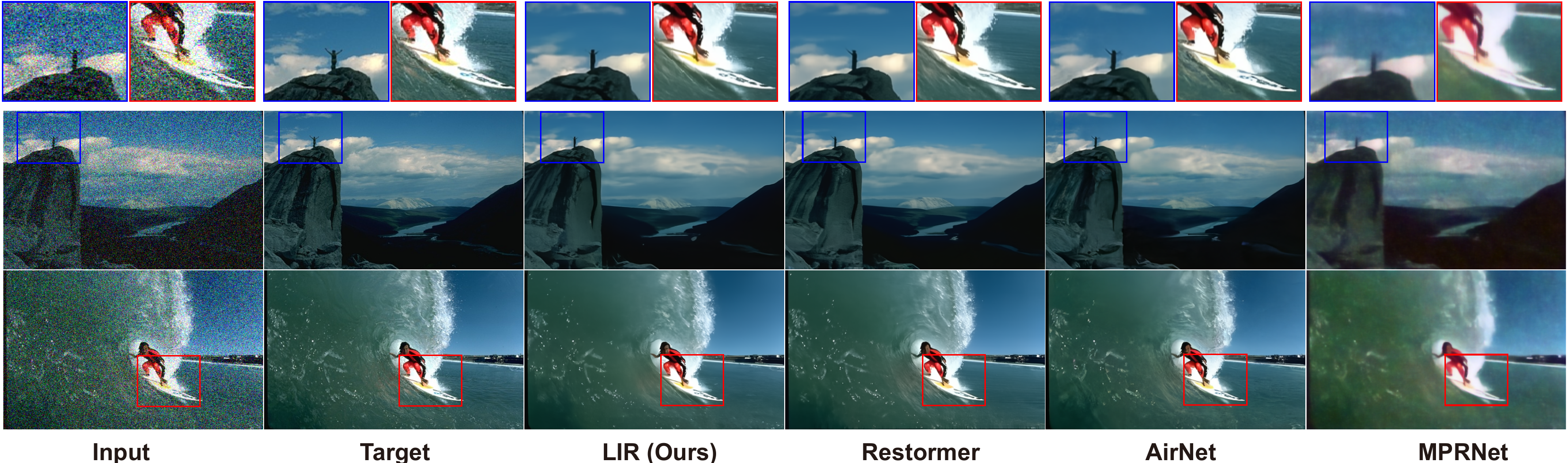}
\caption{The visual comparisons for LIR, Restormer \cite{34zamir2022restormer}, AirNet \cite{54li2022all}, and MPRNet \cite{30zamir2021multi}. LIR produces state-of-the-art visual results.}     
\label{fig:denoise}
\end{figure*}

\subsubsection{Patch Attention}
To give the Attention Block the ability to capture global information, we introduce the Patch Attention (PA) (Figure~\ref{fig:fig4}). In the self-attention part of the transformer, the attention map is calculated using two matrix multiplications and a softmax operation between q, k, and v, which incurs significant computational overhead. Instead, we simplify the attention map computation in our PA by using just one MLP and one element-wise multiplication.

Specifically, we set C, H, and W as the channel, height, and width of the feature map, and $(P_{h}, P_{w})$ as patch size. The $C_{scale}$ is used to control the dimension of the channel. We start by reducing the dimension of the input feature map from $(C, H, W)$ to $(C//C_{scale}, H, W)$ using a $1\times1$ convolution. Next, we divide the feature map into non-overlapping patches and reshape it into $(HW//P_{h}P_{w}, P_{h}P_{w}C//C_{scale})$. Then, we feed this reshaped feature into the MLP to obtain the attention weights for each patch $(HW//P_{h}P_{w}, 1, 1)$. Finally, we perform element-wise multiplication between the obtained attention weights and the input feature map to capture global information. It is important to note that our attention weights have a smaller dimension, resulting in a lower computation scale to self-attention. The $(P_{h}, P_{w})$ in the LIR is set to $(8, 8)$, $(4, 4)$, and $(2, 2)$ respectively, and the $C_{scale}$ is set to 16 and 8, according to the depth of the network.

\begin{figure}[tbp]
\centering
\includegraphics[width=\linewidth]{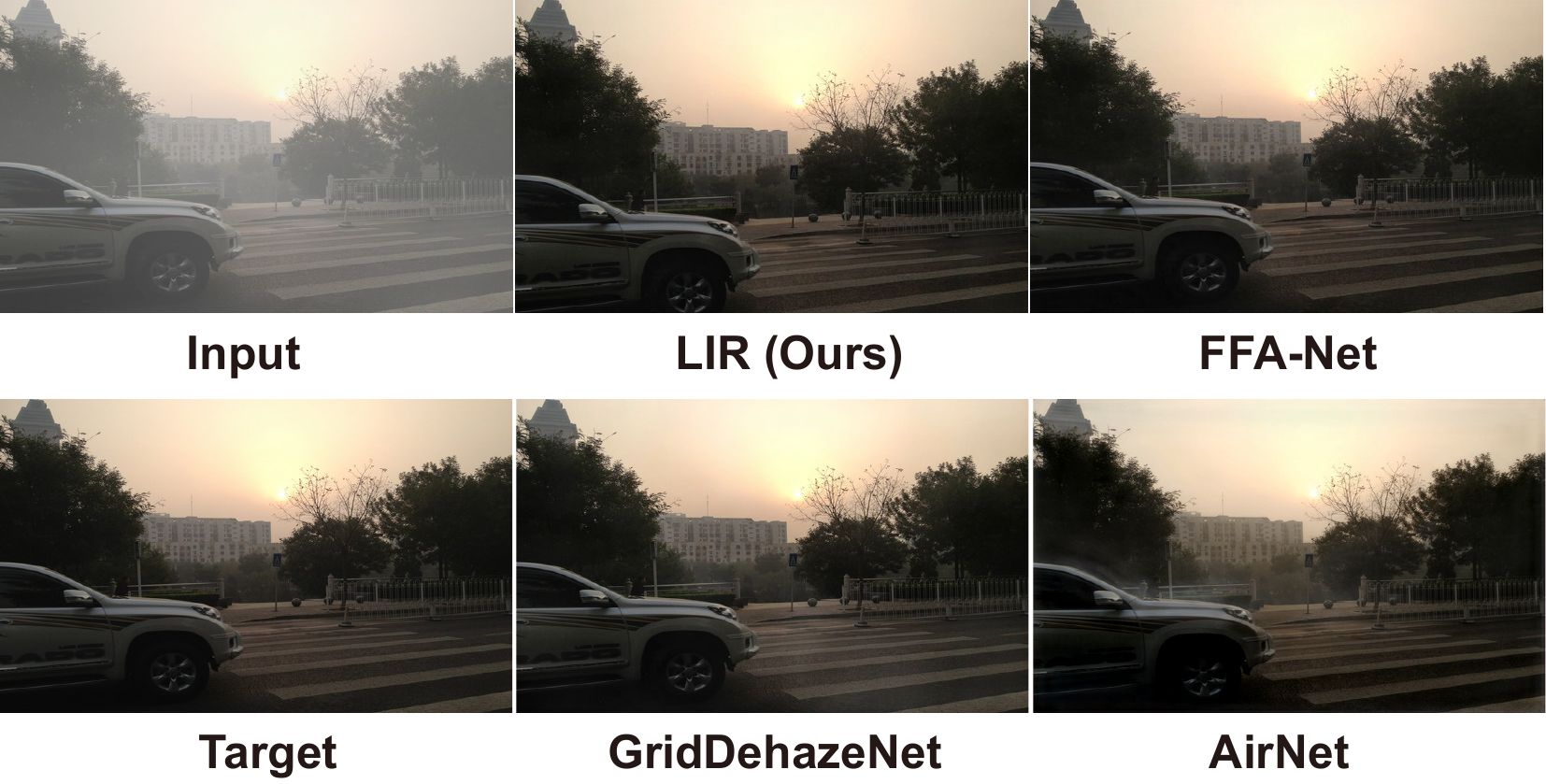}
\caption{The visual comparisons for LIR, FFA-Net \cite{61qin2020ffa}, GridDehazeNet \cite{63liu2019griddehazenet}, and AirNet \cite{54li2022all}. LIR produces better visual results.}    
\label{fig:dehaze}
\end{figure}

\begin{table*}[htbp]
\caption{Image denoising results evaluated by PSNR on CBSD68 and Urban100 datasets. Where the background of light blue for the method that is learned to handle various noise levels.}
\centering
\scalebox{1.1}{
\begin{tabular}{cccccccccc}
\toprule[1.5pt] 
\multirow{2}{*}{Method} & \multirow{2}{*}{Type} & \multicolumn{3}{c}{CBSD68 \cite{bsd68} } & \multicolumn{3}{c}{Urban100 \cite{urban100}} & \multirow{2}{*}{Params {[}M{]}} & \multirow{2}{*}{GFLOPs} \\ \cline{3-8}
                        &                       & $\sigma=15$      & $\sigma=25$      & $\sigma=50$     & $\sigma=15$       & $\sigma=25$      & $\sigma=50$      &                                 &                         \\ \midrule[0.9pt]
FD-GAN \cite{66dong2020fd}                  & GAN                   & \cellcolor[HTML]{DAE8FC}30.25   & \cellcolor[HTML]{DAE8FC}28.81   & \cellcolor[HTML]{DAE8FC}26.43  & -        & -       & -       & 13.98                           & 38.90                   \\
DL \cite{69fan2019general}                       & CNN                   & \cellcolor[HTML]{DAE8FC}33.05   & \cellcolor[HTML]{DAE8FC}30.41   & \cellcolor[HTML]{DAE8FC}26.90  & -        & -       & -       & -                               & -                       \\
MPRNet \cite{30zamir2021multi}                  & CNN                   & \cellcolor[HTML]{DAE8FC}33.54   & \cellcolor[HTML]{DAE8FC}30.89   & \cellcolor[HTML]{DAE8FC}27.56  & -        & -       & -       & 20.13                           & 960.39                  \\
FFDNet \cite{zhang2018ffdnet}                  & CNN                   & \cellcolor[HTML]{DAE8FC}33.87   & \cellcolor[HTML]{DAE8FC}31.21   & \cellcolor[HTML]{DAE8FC}27.96  & \cellcolor[HTML]{DAE8FC}33.83        & \cellcolor[HTML]{DAE8FC}31.40       & \cellcolor[HTML]{DAE8FC}28.05      & -                           & -  \\
DnCNN \cite{zhang2017beyond}                  & CNN                   & \cellcolor[HTML]{DAE8FC}33.90   & \cellcolor[HTML]{DAE8FC}31.24   & \cellcolor[HTML]{DAE8FC}27.95  & \cellcolor[HTML]{DAE8FC}32.98        & \cellcolor[HTML]{DAE8FC}30.81       & \cellcolor[HTML]{DAE8FC}27.59      & -                           & - \\
AirNet \cite{54li2022all}                  & CNN                   & \cellcolor[HTML]{DAE8FC}33.92   & \cellcolor[HTML]{DAE8FC}31.26   & \cellcolor[HTML]{DAE8FC}28.00  & 34.40    & 32.10   & 28.88   & 8.93                            & 169.47                  \\
BRDNet \cite{70tian2020image}                   & CNN                   & 34.10   & 31.43   & 28.16  & 34.42    & 31.99   & 28.56   & -                               & -                       \\
IDR \cite{67zhang2023ingredient}                      & CNN                   & 34.11   & 31.60   & 28.14  & 33.82    & 31.29   & 28.07   & 15.34                           & -                       \\
DRUNet \cite{68zhang2021plug}                  & CNN                   & 34.30   & 31.69   & 28.51  & 34.81    & 32.60   & 29.61   & 32.64                           & 80.78                                     \\
Restormer \cite{46chen2022simple}               & Transformer           & \cellcolor[HTML]{DAE8FC}34.39   & \cellcolor[HTML]{DAE8FC}31.78   & \cellcolor[HTML]{DAE8FC}28.59  & \cellcolor[HTML]{DAE8FC}35.06    & \cellcolor[HTML]{DAE8FC}32.91   & \cellcolor[HTML]{DAE8FC}30.02   & 26.13                           & 92.85                   \\ \midrule[0.9pt]
LIR (Ours)              & CNN                   & \cellcolor[HTML]{DAE8FC}34.31   & \cellcolor[HTML]{DAE8FC}31.69   & \cellcolor[HTML]{DAE8FC}28.50  & \cellcolor[HTML]{DAE8FC}34.81   & \cellcolor[HTML]{DAE8FC}32.59   & \cellcolor[HTML]{DAE8FC}29.57   & 7.31                            & 62.51                   \\ \bottomrule[1.5pt]
\end{tabular}}
\label{denoise}
\end{table*}

\section{Experiment}
\label{experiments}

\subsection{Experiment Settings}
\label{setting}
We set the number of LAA Blocks in the LIR to [3,3,3,3,4], and the width to 48. During the training, we use the $L_{1}$ loss function, AdamW ($\beta_{1}$=0.9, $\beta_{2}$=0.999) optimizer with an initial learning rate of $3e^{-4}$, and set weight decay as $1e^{-4}$. Random image cropping and horizontal flipping are used as data augmentation strategies. Following \cite{34zamir2022restormer}, we use the cosine annealing strategy to gradually decrease the learning rate from $3e^{-4}$ to $1e^{-6}$, and use the progressive learning, where training patch size and batch size pairs are: [($128^{2}$,32), ($160^{2}$,16), ($256^{2}$,8)] at iterations [250K, 250K, 250K] for deblurring, deraining, and denoising. For dehazing, we set [($128^{2}$,32), ($160^{2}$,16)] at iterations [350K, 400K]. The number of parameters and computation for each method are calculated based on the input size of (3,192,192). Specifically, parameters are counted using the PyTorch, and the computation is measured using the thop package.

\subsection{Image Deraining}
\label{Deraining} 
The Rain14000 \cite{fu2017removing} (11,200 training images) and Rain100H \cite{rain100l} (1,254 training images) are used to train LIR, and the Rain100L \cite{rain100l} datasets is used for evaluation. The results are presented in Table~\ref{Rain100L}, where LIR outperforms the DRCNet \cite{49li2022drcnet} and achieves comparable performance to Restormer \cite{34zamir2022restormer} with only a slight 0.02 dB decrease on PSNR with minimal parameters. In addition, our LIR achieves state-of-the-art on SSIM, which significantly outperforms Restormer. It is worth noting that with fewer parameters and computations, LIR significantly outperforms SFNet \cite{51cui2022selective}, the same lightweight design method as ours. Figure~\ref{fig:rain} shows visual comparisons among LIR, Restormer, SFNet, and HINet. It can be demonstrated that LIR produces state-of-the-art visual results, despite LIR having a slightly lower PSNR than Restormer.

\begin{table}[tbp]
\caption{Image dehazing results evaluated by PSNR and SSIM on SOTS outdoor dataset.}
\centering
\scalebox{0.77}{
\begin{tabular}{cccccc}
\toprule[1.5pt] 
\multirow{2}{*}{Method} & \multirow{2}{*}{Type} & \multicolumn{2}{c}{SOTS outdoor \cite{ots_sots}} & \multirow{2}{*}{Params {[}M{]}} & \multirow{2}{*}{GFLOPs} \\ \cline{3-4}
                        &                       & PSNR         & SSIM       &                                 &                         \\ \midrule[0.9pt]
                      
FD-GAN \cite{66dong2020fd}                 & GAN                   & 23.15       & 0.920      & 13.98                           & 38.90                   \\
AirNet \cite{54li2022all}                  & CNN                   & 23.18       & 0.900      & 8.93                            & 169.47                  \\
IDR \cite{67zhang2023ingredient}                      & CNN                   & 25.24       & 0.943      & 15.34                           & -                       \\
GCANet \cite{chen2019gated}                      & CNN                   & 30.23       & 0.980      & -                           & - \\
GridDehazeNet \cite{63liu2019griddehazenet}           & CNN                   & 30.86       & 0.982      & 0.96                            & 13.69                   \\
U2-Former \cite{62ji2021u2}               & Transformer           & 31.10       & 0.976      & -                               & -                       \\
FFA-Net \cite{61qin2020ffa}                 & CNN                   & 33.38       & 0.980      & 4.46                            & 161.74                  \\
MSBDN \cite{dong2020multi}                & CNN                   & 33.67       &  0.985      & -                            & -                   \\
MAXIM-2S \cite{31tu2022maxim}                & MLP                   & 34.19       &  0.985      & 22.5                            & 80.13                   \\\midrule[0.9pt]
LIR (Ours)              & CNN                   & 34.85       &  0.985       & 7.31                            & 62.51                  \\ \bottomrule[1.5pt]
\end{tabular}}
\label{sots}
\end{table}

\begin{table*}[htbp]
\caption{Image deblurring results evaluated by PSNR and SSIM on GoPro and HIDE datasets. }
\centering
\scalebox{1.1}{
\begin{tabular}{cccccccc}
\toprule[1.5pt] 
\multirow{2}{*}{Method} & \multirow{2}{*}{Type} & \multicolumn{2}{c}{GoPro \cite{gopro}} & \multicolumn{2}{c}{HIDE \cite{hide}} & \multirow{2}{*}{Params {[}M{]}} & \multirow{2}{*}{GFLOPs} \\ \cline{3-6}
                        &                       & PSNR         & SSIM        & PSNR        & SSIM       &                         &                        \\ \midrule[0.9pt]
IDR \cite{67zhang2023ingredient}                      & CNN                   & 27.87       & 0.846       & -           & -          & 15.34                   & -                      \\
DeblurGAN-v2 \cite{kupyn2019deblurgan}                      & GAN                   & 29.55       & 0.934       & 26.61           & 0.875          & 59.16                   & 20.12 \\
SRN \cite{tao2018scale}                   & RNN                   & 30.26       & 0.934       & 28.36       & 0.915      & -                   & -                 \\
Gao et al. \cite{gao2019dynamic}                   & CNN                   & 30.90       & 0.935       & 29.11       & 0.913      & -                   & -     \\
DBGAN \cite{74zhang2020deblurring}                   & GAN                   & 31.10       & 0.942       & 28.94       & 0.915      & 11.58                   & 189.71                 \\
MT-RNN \cite{75park2020multi}  & RNN                   & 31.15       & 0.945       & 29.15       & 0.918      & -                       & -                      \\
DMPHN \cite{72zhang2019deep}                   & CNN                   & 31.20       & 0.940       & 29.09       & 0.924      & 21.70                   & -                      \\
Zhe et al. \cite{73jiang2020learning}               & CNN                   & 31.79       &  0.949       & -           & -          & -                       & -                      \\
Suin et al. \cite{71suin2020spatially}              & CNN                   & 31.85       & 0.948       & 29.98       & 0.939      & -                       & -                      \\
SPAIR \cite{53purohit2021spatially}                    & CNN                   & 32.06       & 0.950       &  30.29       &  0.931      & -                       & -                                \\\midrule[0.9pt]
LIR (Ours)              & CNN                   &  32.19       & 0.931        & 30.31       & 0.917       & 7.31                    & 62.51                 \\ \bottomrule[1.5pt]
\end{tabular}}
\label{deblur}
\end{table*}

\subsection{Image Dehazing}
\label{Dehazing}
The OTS \cite{ots_sots} (72,315 training images) is used to train LIR, and the Synthetic Objective Testing Set (SOTS) outdoor \cite{ots_sots} is used for evaluation. In the SOTS outdoor results (Table~\ref{sots}), LIR significantly outperforms both the MAXIM-2S \cite{31tu2022maxim} (MLP-based) and the U2-Former \cite{62ji2021u2} (transformer-based) with improvement in PSNR of 0.66 dB and 3.75 dB. In addition, LIR achieves comparable performance on PSNR and better performance on SSIM to DehazeFormer. Visual comparisons are shown in the Figure~\ref{fig:dehaze}.

\subsection{Image Denoising}
\label{Denoising}
We perform denoising experiments on synthetic benchmark datasets generated by additive white Gaussian noise. The DFWB (DIV2K \cite{div2k}, Flickr2K, WED \cite{wed}, BSD500 \cite{bsd500}) datasets with 8,594 training images are used to train LIR. For evaluation, we use the CBSD68 \cite{bsd68} and Urban100 \cite{urban100} datasets. The background of the light blue color in the Table~\ref{denoise} indicates that the method is trained using a blind denoising setting (random $\sigma$ from [0,50]) and then evaluated under a fixed $\sigma$ (0, 25, or 50). Other methods are trained with a fixed $\sigma$ and evaluated with the same $\sigma$ as training. As shown in the Table~\ref{denoise}, LIR achieves comparable performance to Restormer \cite{34zamir2022restormer} and significantly outperforms all other models on the CBSD68 dataset with a blind denoising setting. Moreover, it is worth noting that with a 77\% reduction in parameters and a 20\% reduction in computations, our LIR achieves comparable performance to DRUNet \cite{68zhang2021plug} on Urban100, while the DRUNet is trained under fixed $\sigma$ setting and our LIR is under random $\sigma$ setting. The visual comparisons between LIR, Restormer, AirNet, and MPRNet are shown in Figure~\ref{fig:denoise}. Although LIR is not as high as Restormer on PSNR, the better visual result is achieved.

\subsection{Image Deblurring}
\label{Deblurring}
The GoPro \cite{gopro} (2,103 training images) dataset is used to train LIR, and the GoPro (1,111 images) and HIDE \cite{hide} (2,025 images) datasets are used for evaluation. Table~\ref{deblur} demonstrates that our LIR significantly outperforms other methods with fewer parameters and computations. Specifically, LIR is 0.13 dB higher than SPAIR \cite{53purohit2021spatially}, and 0.34 dB significantly higher than Suin et al \cite{71suin2020spatially}.

\subsection{Ablation Studies}
\label{Ablation}
We conduct ablation experiments on the deraining task using different types. Type 1 refers to LIR, Type 2 refers to using vanilla global and local residual connections in the model without removing the degradation in residual connections through structural design, Type 3 refers to no Patch Attention module, Type 4 refers to no Adaptive Filter, and Type 5 refers to using vanilla global and local residual connections in the model and not include Patch Attention module and Adaptive Filter. The results in the Table~\ref{ablation} demonstrate that the absence of the Adaptive Filter has the greatest negative impact on LIR, where PSNR is droped by 0.3 dB. Furthermore, it also shown that our proposed Patch Attention module is efficient and computation-friendly, only bringing 0.8 M parameters and 1.81 GFLOPs computations. Although Type 1 only improved performance by 0.09 dB compared to Type 2, structural design for residual connections is the key to the final visual improvement.  

\begin{figure}[htbp]
\centering
\includegraphics[width=\linewidth]{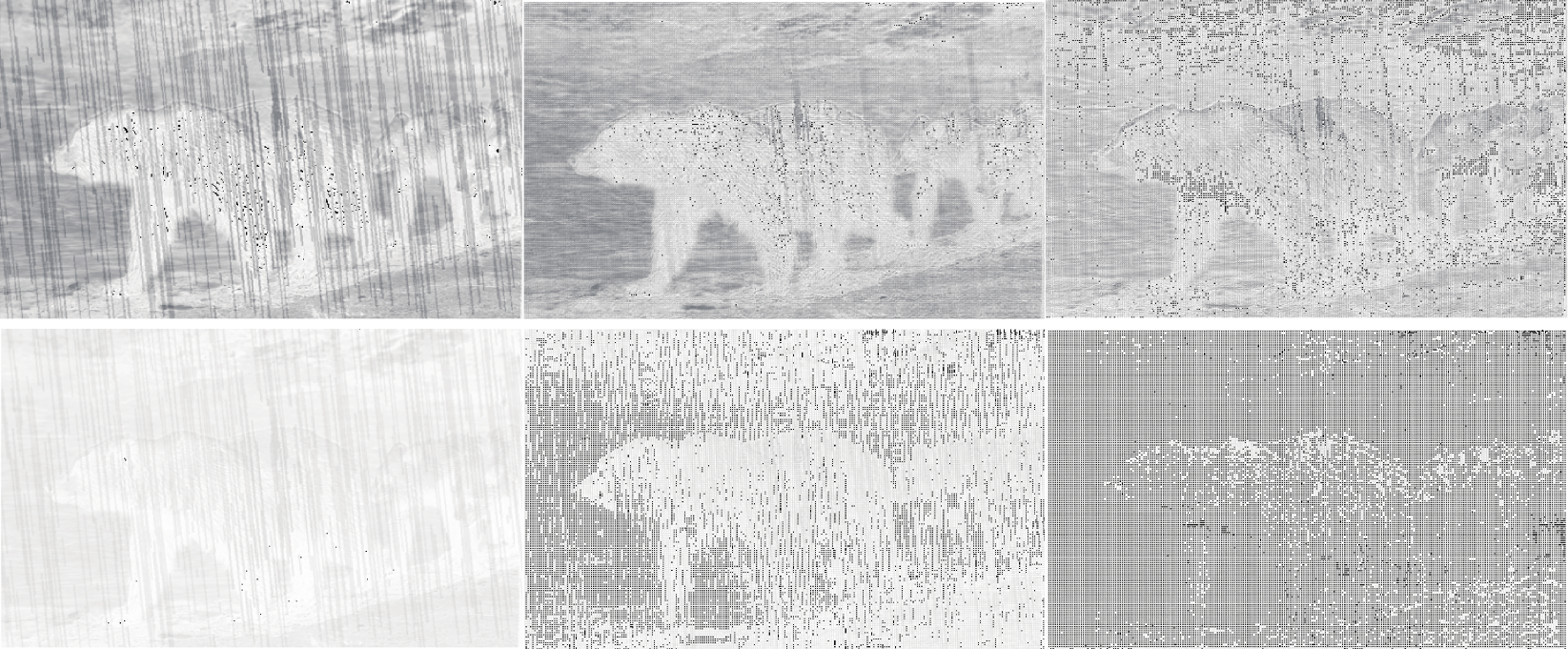}
\caption{The visualization of intermediate feature maps of the Type 1 (top) and the Type 2 (bottom) in the deraining task.}   
\label{fig:visual_ablation1}
\end{figure}

\begin{figure}[htbp]
\centering
\includegraphics[width=\linewidth]{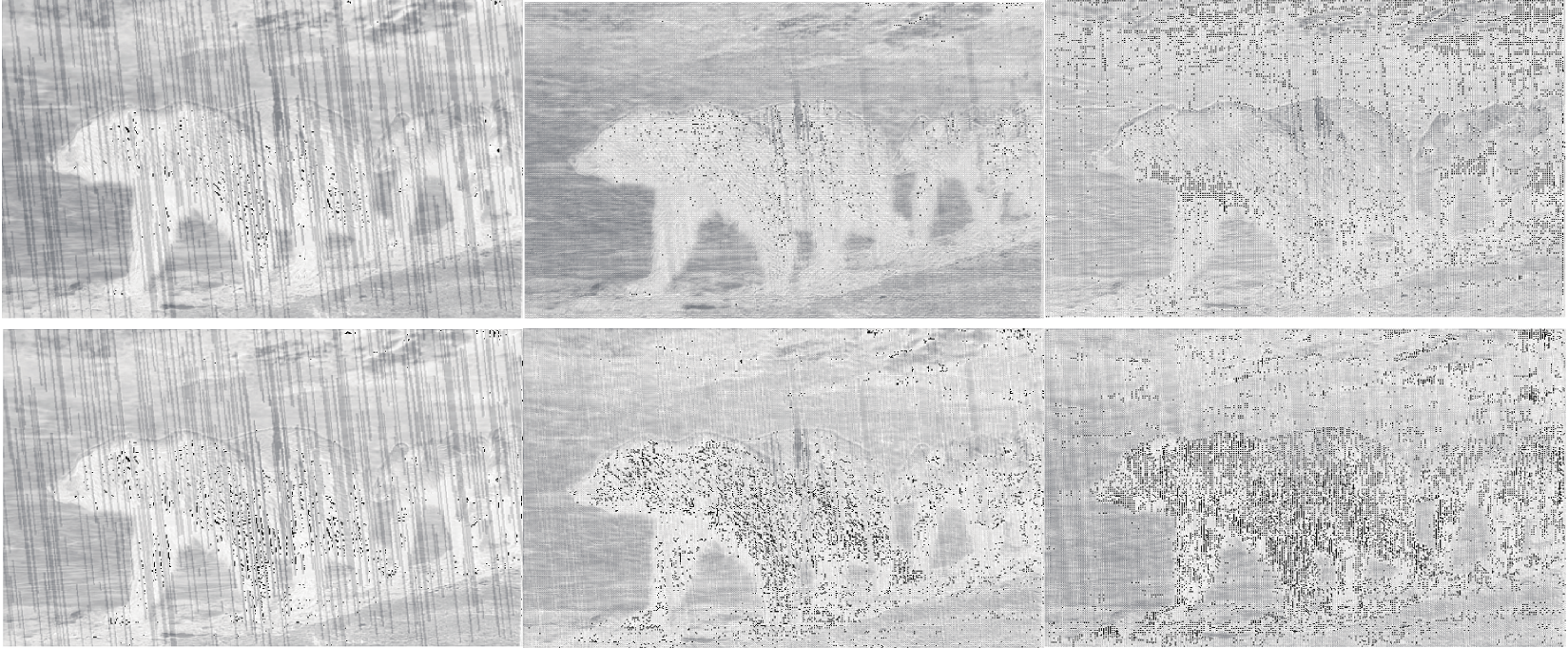}
\caption{The visualization of intermediate feature maps of the Type 1 (top) and the Type 4 (bottom) in the deraining task.}    
\label{fig:visual_ablation2}
\end{figure}

In addition, we present the intermediate feature maps of Type 1 and Type 4 (Figure~\ref{fig:visual_ablation2}). Obviously, LIR with Adaptive Filter has a better visual result than without Adaptive Filter, with less noise and rain trails. We also present the intermediate feature maps of Type 1 and Type 2 (Figure~\ref{fig:visual_ablation1}), where the former has a better visual result than the latter, with fewer rain trails and better contour. More visual results about Type 1, Type 2, and Type 5 please see Supplementary Materials.

\begin{table}[htbp]
\caption{Ablation studies are performed on Rain100L dataset. Where Vanilla Con. (Connections) refers to using vanilla global and local residual connections in the model without removing the degradation in residual connections.}
\centering
\begin{tabular}{cccccc}
\toprule[1.5pt] 
                   & Type 1       & Type 2       & Type 3       & Type 4       & Type 5       \\ \midrule[0.9pt]
Vanilla Con. & \XSolid     & \Checkmark & \XSolid     & \XSolid     & \Checkmark \\
Adaptive Filter    & \Checkmark & \Checkmark & \Checkmark & \XSolid     & \XSolid     \\
Patch Attention    & \Checkmark & \Checkmark & \XSolid     & \Checkmark & \XSolid     \\
Params {[}M{]}     & 7.31         & 6.86         & 7.31         & 6.51         & 5.88         \\
GFLOPs             & 62.51        & 49.15        & 62.51        & 62.15        & 49.01        \\
PSNR               & 38.96        & 38.87        & 38.66        & 38.85        & 37.65        \\ \bottomrule[1.5pt]
\end{tabular}
\label{ablation}
\end{table}
\section{Conclusion and Limitation}
\label{Conclusion}

\subsection{Conclusion}
In this paper, we introduce a Lightweight Image Restoration network called LIR, which aims to efficiently remove degradation. Firstly, the problem that local and global residual connections propagate the degradation throughout the entire network is addressed through the use of the Transpose operation and an ingenious structural design. In addition, we propose the Adaptive Filter to adaptively extract high-frequency information, enhance object contours, and eliminate degradation in various IR tasks. Our Adaptive Filter achieves promising results while requiring minimal parameters and computations, as demonstrated in our ablation studies (Section~\ref{ablation}). To capture global information while keeping small parameters and low computational requirements, we introduce the Patch Attention module by leveraging an MLP and patch embedding strategy. On the deraining task, our LIR achieves the state-of-the-art Structure Similarity Index Measure and comparable performance to state-of-the-art models on Peak Signal-to-Noise Ratio. For denoising, dehazing, and deblurring tasks, LIR also achieves a comparable performance to state-of-the-art models with only about 30\% size.

\subsection{Limitation}
LIR achieves good performance on deraining, denoising and dehazing tasks and better visual results. However, as Figure~\ref{deblur} shows, LIR is less prominent on deblurring tasks than deraining and denoising. We believe that this is because, in the deblurring task, object outlines tend to be stretched out very long. While our Adaptive Filter is very effective when dealing with static object contours, it has limitations when dealing with dynamic contours caused by fast-moving objects.

\bibliographystyle{ACM-Reference-Format}
\bibliography{acmart}

\end{document}